\documentclass[11pt,a4paper]{article}
\usepackage[hyperref]{acl2019}
\usepackage{times}
\usepackage{latexsym}
\usepackage{graphicx} 

\usepackage{url}

\aclfinalcopy 
\def\aclpaperid{914} 

\title{An Investigation of Transfer Learning-Based Sentiment Analysis in Japanese}

\author{Enkhbold Bataa \\
  ExaWizards, Inc. / Tokyo, Japan \\
  \texttt{enkhbold.bataa@exwzd.com} \\\And
  Joshua Wu \\
  ExaWizards, Inc. / Tokyo, Japan \\
  \texttt{joshua.wu@exwzd.com} \\}

\begin{document}
\maketitle
\begin{abstract}
  Text classification approaches have usually required task-specific model architectures and huge labeled datasets. Recently, thanks to the rise of text-based transfer learning techniques, it is possible to pre-train a language model in an unsupervised manner and leverage them to perform effectively on downstream tasks. In this work we focus on Japanese and show the potential use of transfer learning techniques in text classification. Specifically, we perform binary and multi-class sentiment classification on the Rakuten product review and Yahoo movie review datasets. We show that transfer learning-based approaches perform better than task-specific models trained on 3 times as much data. Furthermore, these approaches perform just as well for language modeling pre-trained on $\frac{1}{30}$ of Wikipedia. We release our pre-trained models and code as open source.
\end{abstract}

\section{Introduction}

Sentiment analysis is a well-studied task in the field of natural language processing and information retrieval \cite{sadegh2012opinion,hussein2018survey}. In the past few years, researchers have made significant progress from models that make use of deep learning techniques.\cite{kim2014convolutional,lai2015recurrent,chen2017recurrent, lin2017structured}. However, while there has been significant progress in sentiment analysis for English, not much effort has been invested in analyzing Japanese due to its sparse nature and the dependency on large datasets required by deep learning. Japanese script contains no whitespace, and sentences may be ambiguous such that there are multiple ways to split characters into words, each with a completely different meaning. To see if existing research can make progress in Japanese, we make use of recent transfer learning models such as ELMo \cite{Peters:2018}, ULMFiT \cite{howard2018universal}, and BERT \cite{devlin2018bert} to each pre-train a language model which can then be used to perform downstream tasks. We test the models on binary and multi-class classification.

\begin{figure}[h]
    \centering
    \includegraphics[width=1\linewidth]{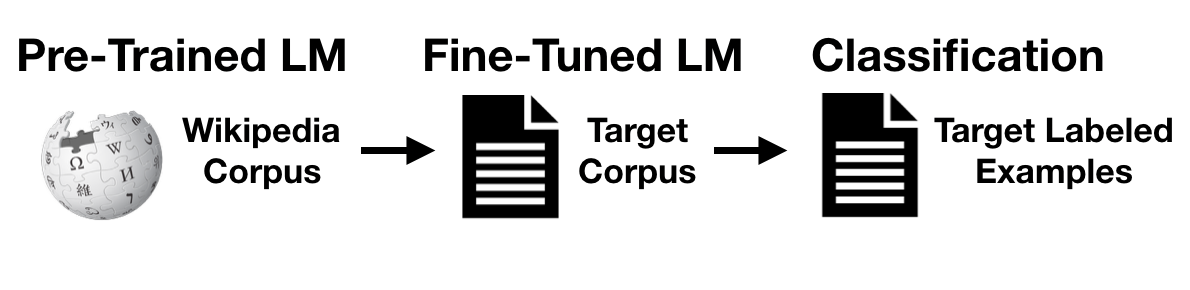}
    \caption{Transfer learning-based text classification. First, we train the LM on a large corpus. Then, we fine-tune it on a target corpus. Finally, we train the classifier using labeled examples.}
    \label{fig:fig_1}
\end{figure}

The training process involves three stages as illustrated in Figure \ref{fig:fig_1}. The basic idea is similar to how fine-tuning ImageNet \cite{deng2009imagenet} helps many computer vision tasks \cite{2016arXiv160808614H}. However, this model does not require labeled data for pre-training. Instead, we pre-train a language model in unsupervised manner and then fine-tune it on a domain-specific dataset to efficiently classify using much less data. This is highly desired since there is a lack of large labeled datasets in practice.

\section{Contributions}
The following are the primary contributions of this
paper:

\begin{itemize}
    \item We experiment ELMo, ULMFiT and BERT on Japanese datasets including binary and 5-class datasets.
    \item We do several ablation studies that are helpful for understanding the effectiveness of transfer learning in Japanese sentiment analysis.
    \item We release our pre-trained models and code\footnote[1]{base.exawizards.com}\footnote[2]{allennlp.org/elmo}
\end{itemize}

\section{Related Work}

Here we briefly review the popular neural embeddings and classification model architectures.  

\subsection{Word Embeddings}

Word embedding is defined as the representation of a word as a dense vector. There have been many neural network implementations, including word2vec \cite{mikolov2013efficient}, fasttext \cite{joulin2016bag} and Glove \cite{pennington2014glove} that embed using a single layer and achieve state-of-the-art performance in various NLP tasks. However, these embeddings are not context-specific: in the phrases "I washed my dish" and "I ate my dish", the word "dish" refers to different things but are still represented by the same embedding.

\subsection{Contextualized Word Embeddings}

Instead of fixed vector embeddings, Cove \cite{mccann2017learned} uses a machine translation model to embed each word within the context of its sentence. The model includes a bidirectional LSTM encoder and a unidirectional LSTM decoder with attention, and only the encoder is used for downstream task-specific models. However, pre-training is limited by the availability of parallel corpora. (e.g. English-French)

ELMo, short for Embeddings from Language Model \cite{Peters:2018} overcomes this issue by taking advantage of large monolingual data in an unsupervised way. The core foundation of ELMo is the bidirectional language model which learns to predict the probability of a target word given a sentence by combining forward and backward language models. ELMo also requires task-specific models for downstream tasks.

\citet{howard2018universal} proposed a single-model architecture, ULMFiT, that can be used in both pre-training and task-specific fine-tuning. They use novel techniques such as discriminative fine-tuning and slanted triangular learning rates for stable fine-tuning. OpenAI extended the idea by introducing GPT, a multi-layer transformer decoder \cite{radford2018improving}. While ELMo uses shallow concatenation of forward and backward language models, ULMFiT and OpenAI GPT are unidirectional. 

\citeauthor{devlin2018bert} argues that this limits the power of pre-trained representations by not incorporating bidirectional context, crucial for word-level tasks such as question answering. They proposed a multi-layer transformer encoder-based model, BERT, trained on masked language modeling (MLM) and next sentence prediction (NSP) tasks. MLM allows bidirectional training by randomly masking 15\% of words in each sentence in order to predict them, and NSP helps tasks such as question answering by predicting the order of two sentences.

\subsection{Text Classification}
\label{abl:text_classification}
Many models have been invented for English text classification, including KimCNN \cite{kim2014convolutional}, LSTM \cite{chen2017recurrent}, Attention \cite{chen2017recurrent}, RCNN \cite{lai2015recurrent}, etc. However, not much has been done for Japanese. To the best of our knowledge, the current state-of-the-art for Japanese text classification uses shallow (context-free) word embeddings for text classification \cite{Zhang_japanesesentiment,nio2018japanese}. \citet{2018arXiv181007653S} proposed the Super Characters method that converts sentence classification into image classification by projecting text into images. 

\citet{2017arXiv170802657Z} did an extensive study of different ways of encoding Chinese/Japanese/Korean (CJK) and English languages, covering 14 datasets and 473 combinations of different encodings including one-hot, character glyphs, and embeddings and linear, fasttext and CNN models. 

This paper investigates transfer learning-based methods for sentiment analysis that is comparable to above mentioned models including \citet{2017arXiv170802657Z} and \citet{2018arXiv181007653S} for the Japanese language. 

\section{Dataset}

Our work is based on the Japanese Rakuten product review binary and 5 class datasets, provided in \citet{2017arXiv170802657Z} and an Yahoo movie review dataset.\footnote[3]{github.com/dennybritz/sentiment-analysis} Table \ref{tab:table_one}  provides a summary. The Rakuten dataset is used for comparison purposes, while the Yahoo dataset is used for ablation studies due to its smaller size. For the Rakuten dataset, 80\% is used for training, 20\% for validation, and the test set is taken from \citet{2017arXiv170802657Z}; for the Yahoo dataset, 60\% is used for training, 20\% for validation, and 20\% for testing. We used the Japanese Wikipedia\footnote[4]{dumps.wikimedia.org/} for pre-training the language model for all models so that comparison would be fair. 

\begin{table}
\centering
\small
    \begin{tabular}{c c c c} 
 \hline
 Dataset & Classes & Train & Test \\ [0.1ex] 
 \hline
 Rakuten full & 5 & 4,000,000 & 500,000 \\ 
 Rakuten binary & 2 & 3,400,000 & 400,000 \\
 Yahoo binary & 2 & 30545 & 7637 \\ 
\end{tabular}
    \caption{Datasets}
    \label{tab:table_one}
\end{table}

\section{Training}

\subsection{Pre-Training Language Model}

Pre-training a language model is the most expensive part but we train it only once and fine-tune on a target task. We used 1 NVIDIA Quadro GV100 for training ULMFiT and 4 NVIDIA Tesla V100s for ELMo. Text extraction done by  WikiExtractor\footnote[5]{github.com/attardi/wikiextractor}, then tokenized by Mecab\footnote[6]{taku910.github.io/mecab/} with IPADIC neologism dictionary\footnote[7]{github.com/neologd/mecab-ipadic-neologd}. We didn't use the BERT multilingual model\footnote[8]{github.com/google-research/bert/blob/master/multilingual.md} due to its incompatible treatment of Japanese: it does not account for \emph{okurigana} (verb conjugations) and diacritic signs which completely change the represented word (e.g. \emph{aisu} "to love" vs. \emph{aizu} "signal").\footnote[9]{github.com/google-research/bert/issues/133}\footnotemark[10] Instead, we use the pre-trained BERT$_{BASE}$ model by \citet{bertjapanese} which has been trained for 1 million steps with sequence length of 128 and 400 thousand additional steps with sequence length of 512. It used the SentencePiece subword tokenizer\cite{kudo2018sentencepiece} for tokenization. The models trained with the most frequent 32000 tokens or subwords.

\subsection{Fine-Tuning}

We use a biattentive classification network (BCN) from \citet{mccann2017learned} with ELMo as it is known to be state-of-the-art\footnotemark[11] on SST \cite{socher2013recursive} datasets. For fine-tuning all models on a target task, we follow the same parameters that were used in the original implementation.\footnotemark[12]\footnotemark[13]\footnotemark[14] And the same hardware used for pre-training ULMFiT and ELMo in fine-tuning. For BERT, we used single v2.8 TPU.\footnotemark[15]

\section{Results}

In this section, we compare the results of ELMo+BCN, ULMFiT, and BERT with models reported in \citet{2017arXiv170802657Z} and other previous state-of-the-art models we mentioned in \ref{abl:text_classification}. Note that none of the source LM is fine-tuned on a target dataset. Results with these models fine-tuned on target corpora are included in Section \ref{abl:domain_adaptation}.

\subsection{Rakuten Datasets}

We trained ELMo+BCN and ULMFiT on the Rakuten datasets for 10 epochs each and selected the one that performed best. Since BERT fine-tunes all of its layers, we only train for 3 epochs as suggested by \citet{devlin2018bert}. Results are presented in Table \ref{tab:table_two}. All transfer learning-based methods outperform previous methods on both datasets, showing that these methods still work well without being fine-tuned on target corpora. 
\begin{table}
\small
\centering
    \begin{tabular}{c c c} 
 \hline
 Model & Rakuten Binary & Rakuten Full \\ [0.1ex] 
 \hline
 GlyphNet & 8.55 & 48.97\\ 
 OnehotNet & 5.93 & 45.1\\
 EmbedNet & 6.07 & 45.2\\
 Linear Model & 6.63 & 45.26\\
 Fasttext & 5.45 & 43.27\\
 Super Character & 5.15 & 42.30\\
 \hline
 BCN+ELMo & 4.77 & 42.95\\
 \textbf{ULMFiT} & \textbf{4.45} & 41.39\\
 \textbf{BERT$_{BASE}$} & 4.68 & \textbf{40.68}\\
 \hline
\end{tabular}
    \caption{Rakuten test results, in error percentages. Best results from other models (GlyphNet to Super Character) obtained from \citet{2017arXiv170802657Z} and \citet{2018arXiv181007653S} }
    \label{tab:table_two}
\end{table}

\footnotetext[10]{github.com/google-research/bert/issues/130}
\footnotetext[11]{nlpprogress.com/english/sentiment\_analysis.html}
\footnotetext[12]{github.com/fastai/fastai}
\footnotetext[13]{github.com/allenai/allennlp}
\footnotetext[14]{github.com/google-research/bert\#fine-tuning-with-bert}
\footnotetext[15]{cloud.google.com/tpu/}

\subsection{Yahoo movie review dataset}

The Yahoo dataset is approximately 112 times smaller than the Rakuten binary dataset. We believe that this dataset better represents real life/practical situations. For establishing a baseline, we trained a simple one-layer RNN and an LSTM with one linear layer on top for classification, as well as convolutional, self-attention, and hybrid state-of-the-art models we mentioned in Section \ref{abl:text_classification} for comparison. Results shown on Table \ref{tab:table_three}. Similar to rakuten datasets, transfer-learning based methods works better.

\begin{table}
\centering
\small
    \begin{tabular}{c c} 
 \hline
 Model & Yahoo Binary \\ [0.1ex] 
 \hline
 RNN Baseline & 35.29\\ 
 LSTM Baseline & 32.41\\
 KimCNN \citet{kim2014convolutional} & 14.25\\
 Self Attention \citet{lin2017structured} & 13.16\\
 RCNN \citet{lai2015recurrent} & 12.67\\
 \hline
 BCN+ELMo & 10.24\\
 ULMFiT & 12.20\\
 \textbf{BERT$_{BASE}$} &  \textbf{8.42}\\
 \hline
\end{tabular}
    \caption{Yahoo test results, in error percentages.}
    \label{tab:table_three}
\end{table}

\section{Ablation Study}
\label{abl:ablation_study}

\subsection{Domain Adaptation}
\label{abl:domain_adaptation}
Pre-trained language models are usually trained with general corpuses such as Wikipedia. However, the target domain corpus distribution is usually different(movie or product review in our case). Therefore, we fine-tune each source language model on the target corpus (without labels) for a few iterations before training each classifier. The results in Table \ref{tab:table_four} shows that fine-tuning ULMFiT improves the performance on all datasets while ELMo and BERT shows varied results. We believe that the huge performance improvement of ULMFiT is due to the discriminative fine-tuning and  slanted triangular learning rates \cite{howard2018universal} that are used during the domain adaptation process.

\begin{table}
\small
\centering
    \begin{tabular}{c c c c} 
 \hline
 Model & Rak B & Rak F & Yahoo B \\ [0.1ex] 
 \hline
 BCN+ELMo & 4.77 & 42.95 & 10.24\\
 ULMFiT & 4.45 & 41.39 & 12.20\\
 BERT$_{BASE}$ & 4.68 & 40.68 & 8.42\\
 \hline
 BCN+ELMo$\ast$ & 4.65 & 43.12 & 8.76\\
 ULMFiT$\ast$ &\textbf{4.18} & 41.05 & \textbf{8.52}\\
 BERT [10K steps]$\ast$ & 4.94 & \textbf{40.52} & 10.14\\
 BERT [50K steps]$\ast$ & 5.52 & 40.57 & -\\
 \hline
\end{tabular}
    \caption{Domain adapted results. ULMFiT$\ast$ and ELMo$\ast$ are trained for 5 epochs, while BERT$\ast$ is trained for 10K and 50K steps.}
    \label{tab:table_four}
\end{table}

\subsection{Low-Shot Learning}

Low-shot learning refers to the practice of feeding a model with a small amount of training data, contrary to the normal practice of using a large amount of data. We chose the Yahoo dataset for this experiment due to its small size. Experimental results in Table \ref{tab:table_five} show that, with only $\frac{1}{3}$ of the total dataset, ULMFiT and BERT perform better than task-specific models, while BCN+ELMo shows a comparable result. Clearly, this shows that the models have learned significantly during the transfer learning process.

\begin{table}
\small
\centering
    \begin{tabular}{c c} 
 \hline
 Model & Yahoo Binary \\ [0.1ex] 
 \hline
 RNN Baseline & 35.29\\ 
 LSTM Baseline & 32.41\\
 KimCNN \citet{kim2014convolutional} & 14.25\\
 Self-Attention \citet{lin2017structured} & 13.16\\
 RCNN \citet{lai2015recurrent} & 12.67\\
 \hline
 BCN+ELMo [$\frac{1}{3}$] & 13.51\\
 ULMFiT Adapted [$\frac{1}{3}$] & 10.62\\
 \textbf{BERT$_{BASE}$} [$\frac{1}{3}$] &  \textbf{10.14}\\
 \hline
\end{tabular}
    \caption{Low-shot learning results for the Yahoo dataset, in error percentages. Transfer learning-based methods are trained on $\frac{1}{3}$ of the total dataset, while the other models are trained on the whole dataset.}
    \label{tab:table_five}
\end{table}

\subsection{Size of Pre-Training Corpus}

We also investigate whether the size of the source language model affects the sentiment analysis performance on the Yahoo dataset. This is especially important for low-resource languages that do not usually have large amounts of data available for training.
We used the ja.text8\footnote[16]{github.com/Hironsan/ja.text8} small text corpus (100MB) from the Japanese Wikipedia to compare with the whole Wikipedia (2.9GB) used in our previous experiments. Table \ref{tab:table_six} shows slightly lower performance for BCN+ELMo and ULMFiT  while BERT performed much worse. Thus, for effective sentiment analysis, a large corpus is required for pre-training BERT. 

\begin{table}
\small
\centering
    \begin{tabular}{c c} 
 \hline
 Model & Yahoo Binary \\ [0.1ex] 
 \hline
 BCN+ELMo & 10.24\\
 ULMFiT & 12.20\\
 ULMFiT Adapted &8.52\\
 \textbf{BERT$_{BASE}$} &  \textbf{8.42}\\
 \hline
 BCN+ELMo [100MB] & 10.32\\
 ULMFiT Adapted [100MB] & 8.57\\
 BERT$_{BASE}$ [100MB]  & 14.26\\
 \hline
\end{tabular}
    \caption{Comparison of results using large and small corpora. The small corpus is uniformly sampled from the Japanese Wikipedia (100MB). The large corpus is the entire Japanese Wikipedia (2.9GB).}
    \label{tab:table_six}
\end{table}

\subsection{Parameter Updating Methods}
In its original implementation, when BERT is fine-tuned, all of its layers are trained. This is quite different from fine-tuning ELMo, where its layers are frozen and only task-specific models (BCN in our case) are updated. We experiment with the opposite case for both models and list the results on Table \ref{tab:table_seven}

\begin{itemize} 
\item
\textbf{BERT as a feature extractor} Pre-trained BERT weights are used for initialization and will not be changed. The hidden state associated to the first character of the input is pooled and provided to a linear layer that sits on top. This way, BERT is computationally much cheaper and faster. Result shows that using BERT as a feature extractor shows competitive performance.
\item
\textbf{Unfreezing ELMo} Pre-trained ELMo weights are used for initialization as well; however, weights are changed along with BCN layers. This experiment allows us to compare the performance of freezing/unfreezing ELMo layers. Table \ref{tab:table_seven} shows that fine-tuning ELMo improves performance, comparable to BERT. 
\end{itemize}

\begin{table}
\small
\centering
    \begin{tabular}{c c} 
 \hline
 Model & Yahoo Binary \\ [0.1ex] 
 \hline
 BCN+ELMo & 10.24\\
 BCN+ELMo unfreeze & 8.65\\
 BERT$_{BASE}$ & 8.42\\
 BERT$_{BASE}$ freeze &  10.68\\
 \hline
\end{tabular}
    \caption{Results from different parameter updating strategies. BCN+ELMo and BERT$_{BASE}$ are original implementations. BCN+ELMo unfreeze shows experimental results of fine-tuning both BCN and ELMo layers on target dataset while BERT$_{BASE}$ freeze is where BERT$_{BASE}$ layers are frozen and only classifier layer fine-tuned on target dataset}
    \label{tab:table_seven}
\end{table}

\section{Conclusion}
Our work showed the possibility of using transfer learning techniques for addressing sentiment classification for the Japanese language. We draw following conclusions for future researchers in Japanese doing transfer learning for sentiment analysis task based on experiments we did in Rakuten product review and Yahoo movie review datasets:
\begin{enumerate}
    \item Adapting domain for BERT likely will not yield good results when the task is binary classification. For all other cases, domain adaptation performs just as well or better.
    \item ELMo and ULMFiT perform well even when trained on a small subset of the language model.
    \item Fune-tuning both ELMo and BCN layers on a target task improves the performance.
\end{enumerate}

\section{Discussion and Future Considerations}

This research is a work in progress and will be regularly updated with new benchmarks and baselines. We showed that with only $\frac{1}{3}$ of the total dataset, transfer learning approaches perform better than previous state-of-the-art models. ELMo and ULMFiT perform just as well trained on small corpora, but BERT performs worse since it is designed to be trained on MSM and NSP tasks. Finally, domain adaptation always improves the performance of ULMFiT. We believe that our ablation study and the release of pre-trained models will be particularly useful in Japanese text classification. It is important to note that we did not perform K-fold validation due to their high computational cost. In the future, we will investigate other NLP tasks such as named entity recognition (NER), question answering (QA) and aspect-based sentiment analysis (ABSA) \cite{pontiki2016semeval} to see whether results we saw in sentiment analysis is consistent across these tasks. We hope that our experimental results inspire future research dedicated to Japanese. 

\bibliography{acl2019}
\bibliographystyle{acl_natbib}

\end{document}